\documentclass[conference]{IEEEtran}
\IEEEoverridecommandlockouts
\usepackage{cite}
\usepackage{amsmath,amssymb,amsfonts}
\usepackage{algorithmic}
\usepackage{graphicx}
\usepackage{textcomp}
\usepackage{xcolor}
\usepackage{subfigure}
\usepackage{setspace}
\usepackage{array}
\usepackage{geometry}
\def\BibTeX{{\rm B\kern-.05em{\sc i\kern-.025em b}\kern-.08emT\kern-.1667em\lower.7ex\hbox{E}\kern-.125emX}}
\usepackage[colorlinks,linkcolor=blue,anchorcolor=blue,citecolor=blue]{hyperref} 
\usepackage[colorinlistoftodos]{todonotes}  
\newcommand{\etal}{\emph{et al.}}           

\renewcommand{\baselinestretch}{0.94}
\begin{document}

\title{To What Extent Does Downsampling, Compression, and Data Scarcity Impact Renal Image Analysis?}

\author{\IEEEauthorblockN{
			Can Peng\IEEEauthorrefmark{1},
			Kun Zhao\IEEEauthorrefmark{1},
			Arnold Wiliem\IEEEauthorrefmark{1},
			Teng Zhang\IEEEauthorrefmark{1},
			Peter Hobson\IEEEauthorrefmark{2},
			Anthony Jennings\IEEEauthorrefmark{2},
			Brian C. Lovell\IEEEauthorrefmark{1}}
		\IEEEauthorblockA{\IEEEauthorrefmark{1} School of ITEE, The University of Queensland, Brisbane, QLD, 		Australia\\
		\IEEEauthorrefmark{2}Sullivan Nicolaides Pathology, Australia}}
\maketitle

\begin{abstract}
The condition of the Glomeruli, or filter sacks, in renal Direct Immunofluorescence (DIF) specimens is a critical indicator for diagnosing kidney diseases. A digital pathology system which digitizes a glass histology slide into a Whole Slide Image (WSI) and then automatically detects and zooms in on the glomeruli with a higher magnification objective will be extremely helpful for pathologists. In this paper, using glomerulus detection as the study case, we provide analysis and observations on several important issues to help with the development of Computer Aided Diagnostic (CAD) systems to process WSIs. Large image resolution, large file size, and data scarcity are always challenging to deal with. To this end, we first examine image downsampling rates in terms of their effect on detection accuracy. Second, we examine the impact of image compression. Third, we examine the relationship between the size of the training set and detection accuracy. To understand the above issues, experiments are performed on the state-of-the-art detectors: Faster R-CNN, R-FCN, Mask R-CNN and SSD. Critical findings are observed: (1) The best balance between detection accuracy, detection speed and file size is achieved at 8 times downsampling captured with a $40\times$ objective; (2) compression which reduces the file size dramatically, does not necessarily have an adverse effect on overall accuracy; (3) reducing the amount of training data to some extents causes a drop in precision but has a negligible impact on the recall; (4) in most cases, Faster R-CNN achieves the best accuracy in the glomerulus detection task. We show that the image file size of $40\times$ WSI images can be reduced by a factor of over 6000 with negligible loss of glomerulus detection accuracy.
\end{abstract}

\section{Introduction}
The introduction of the kidney biopsy is one of the major events in the history of nephrology \cite{agarwal2013basics}. The kidney biopsy helps diagnose diseases such as glomerulonephritis and glomerulosclerosis \cite{agarwal2013basics}. Direct Immunofluorescence (DIF) is usually used as the gold standard for immunohistochemical evaluation of renal biopsy specimens \cite{molne2005immunoperoxidase}. Traditionally, specimen diagnosis is performed manually by pathologists with microscopes. This process is subjective, time-inefficient and labour-intensive \cite{zhao2018dgdi}. In addition, due to fluorescence bleaching, DIF slides can only be stored and viewed for a limited period. To solve these problems, many works such as \cite{samak2015optimization} have developed systems that can digitize these slices for permanent recording and consequently allow computer-aided analysis.

\begin{figure}
	\centering
	\includegraphics[height=5cm,width=8.8cm]{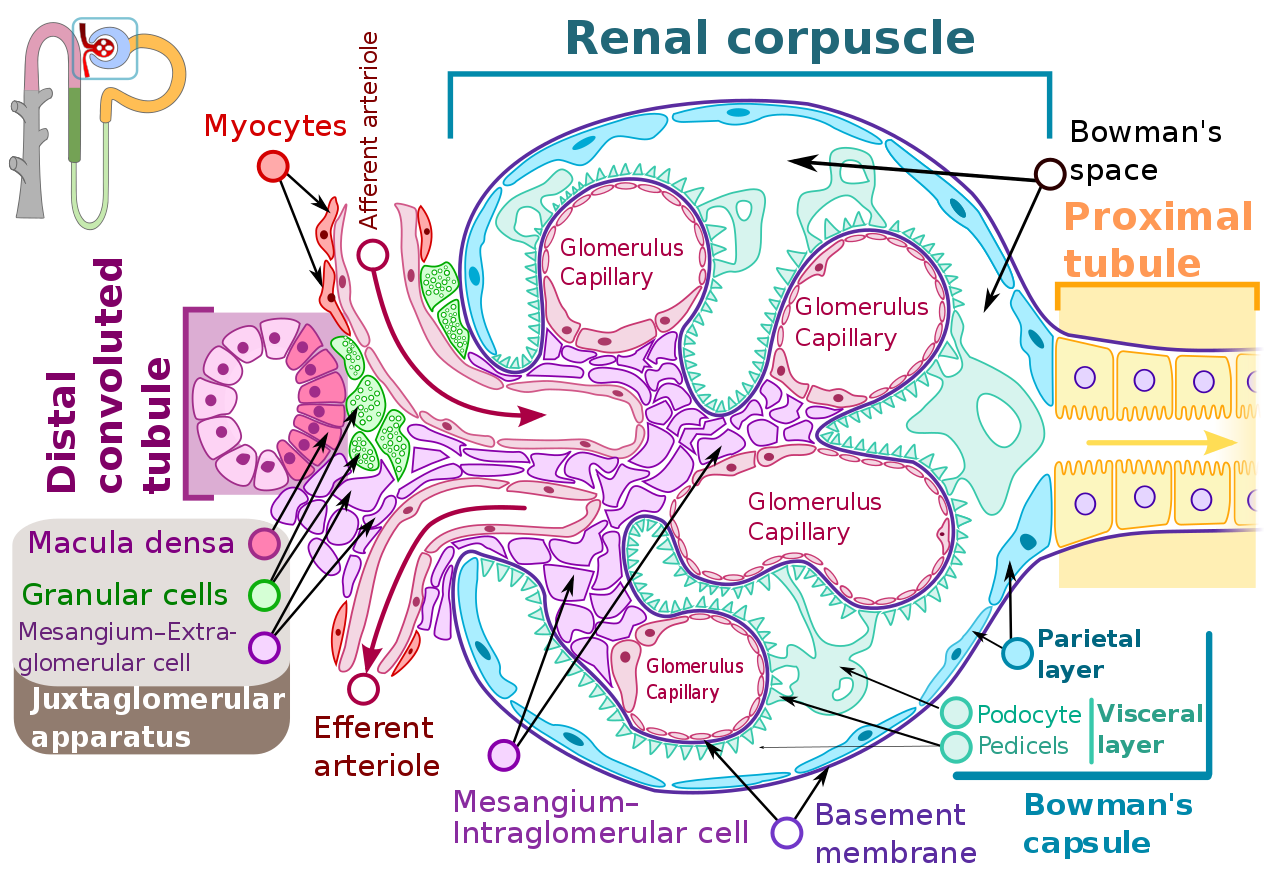}
	\caption{Diagram of the renal corpuscle structure of the glomerulus
(\href{http://comenius.susqu.edu/biol/202/animals/terms/figures/glomerulus.gif}
{this image} by
\href{https://commons.wikimedia.org/wiki/User:M.Komorniczak}{M. Komorniczak} is licensed under
\href{https://creativecommons.org/licenses/by-sa/4.0/deed.en}{CC BY-SA 4.0})}
	\label{fig:glomerulus}
\end{figure}

\textbf{The challenges} of DIF WSI renal analysis include large image resolution, large file size, and data scarcity. The DIF WSIs are extremely large with file size up to tens of gigabytes with resolution up to gigapixels. It is impossible to directly feed such large images to Computer Aided Diagnostic (CAD) systems based on Convolutional Neural Networks (CNNs). Moreover, hospitals would like to store patients’ diagnostic results for future review. However, storing such large DIF files will be very expensive and hard to manage. Thus, pre-processing is a critical step to automate the WSI analysis task and it is vital to explore to what extent the pre-processing methods will affect the analysis. Furthermore, compared to many general images, renal biopsy images are significantly more costly and invasive to obtain as they involve surgery on patients. Lack of training data will affect the CAD system's performance.

\begin{figure}
	\centering
	\includegraphics[height=5cm,width=8.8cm]{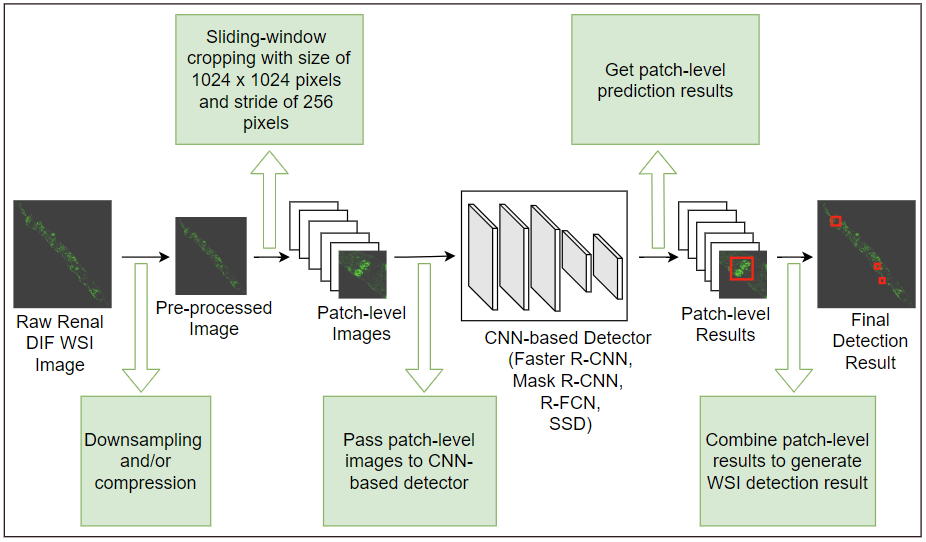}
	\caption{The main procedure for glomerulus detection on the renal WSI --- The raw renal DIF WSI is first pre-processed by downsampling and compressing to reduce both image resolution and file size. Then, the pre-processed image is cropped into patches of suitable size through a sliding-window method. After that, the patches are fed into the CNN detector to get detection result. Finally, the patch-level results are combined to generate the final WSI detection result.}
	\label{fig:framework}
\end{figure}

\begin{figure}
	\centering
	\includegraphics[height=5cm,width=8.8cm]{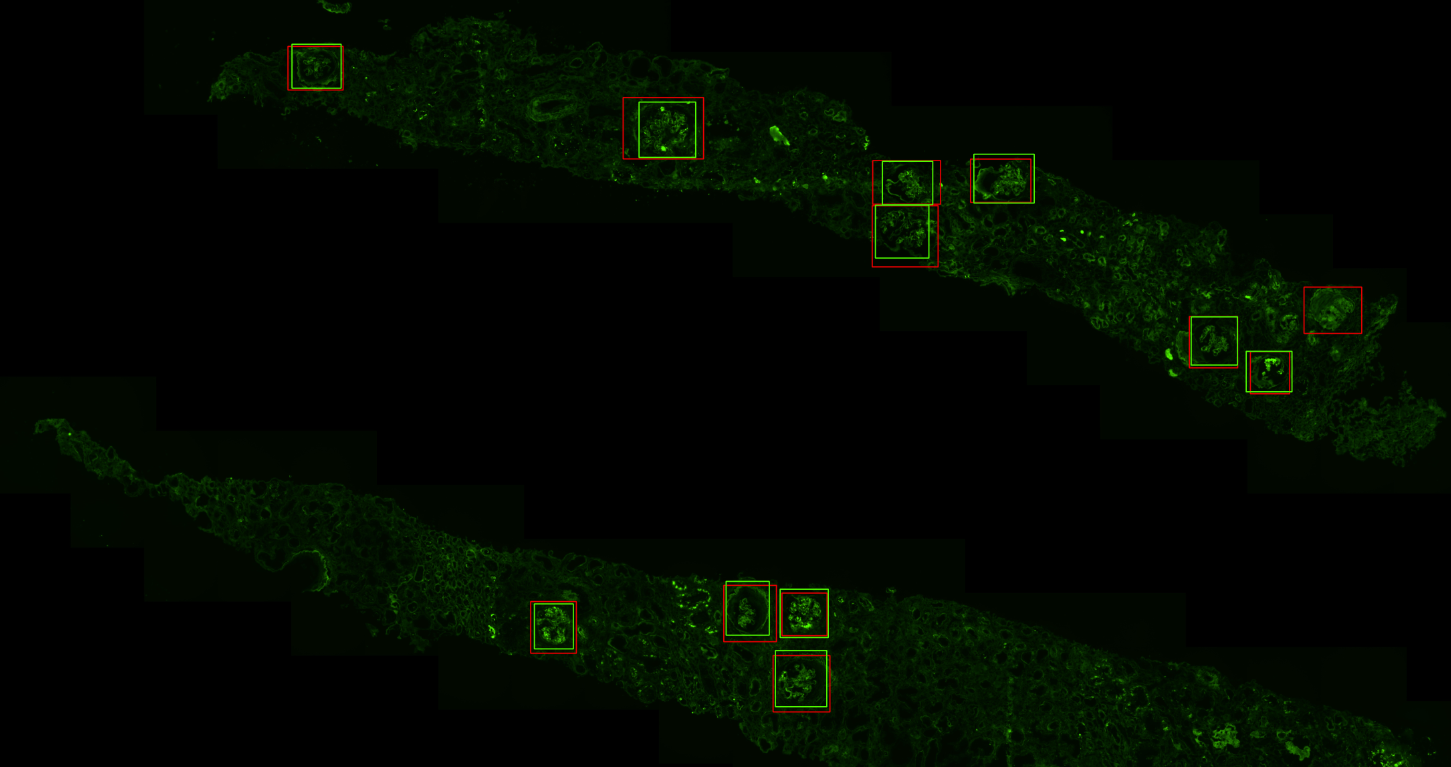}
	\caption{An example of glomerulus detection results for a renal DIF WSI. The red boxes are the ground truth bounding boxes and the green boxes are the predicted bounding boxes.}
	\label{fig:wsi_detection}
\end{figure}
One way to address the above issues is to first detect areas/objects of interest used by the pathologists to make the correct diagnoses. This can also reduce the time in viewing the WSIs as the system can bring up the locations of detected area/object of interests. Note that, in DIF WSIs analysis, glomeruli are the primary objects of interest. Thus, we confine ourselves to study the effect of the pre-processing steps and amount of training data on the glomerulus detection problem. Figure~\ref{fig:glomerulus} shows the renal corpuscle structure of the glomerulus. For our experiments, the image data were captured using a microscope camera with $40\times$ magnification. Figure~\ref{fig:framework} illustrates the whole detection procedure.

\textbf{The findings} of the study are as follows:
\begin{itemize}
	\item With a fixed patch size (1024 $\times$ 1024 pixels), downsampling affects the detection accuracy. If the image is not downsampled, the image patch will only contain a whole or partial glomerulus. Training a model under these conditions increases the false positive rate as background information is insufficient. On the other hand, significantly reducing the resolution of the image increases the false negative rate as the glomeruli become small and ambiguous.	
	\item Applying JPEG compression to the raw renal DIF WSI does not affect detection performance significantly. However, the optimal compression for each detection model can be different. This finding may help in addressing the data storage issues.
	
	\item Reducing the amount of training data to some extents causes a drop in precision but has a negligible impact on the recall. The rate of mean Average Precision (mAP) performance drop is significantly smaller than the rate of data reduction. For instance, reducing the training data by 20\%  only reduces Faster R-CNN detection performance from 0.781 mAP to 0.734 mAP at an Intersection of Union (IoU) threshold of 0.5. Note that, this finding does not apply to SSD\cite{liu2016ssd} which is also the worst performing method in the study.
	
	\item Of the four object detectors: Faster R-CNN \cite{ren2015faster}, Mask R-CNN\cite{he2017mask}, R-FCN\cite{dai2016r} and SSD\cite{liu2016ssd}, Faster R-CNN has the best performance of 0.781 mAP at an IoU of 0.5.
\end{itemize}

We continue our paper as follows. In Section~\ref{rl}, we discuss the related work.
Section~\ref{ex} presents the experimental setup and protocols. Section~\ref{re} presents and discusses the results followed by Section~\ref{conclu} presenting the conclusions.

\section{Related Work}
\label{rl}

We first discuss common pre-processing methods in renal WSI analysis followed by discussions on recent state-of-the-art object detection methods.

\subsection{Renal WSI Pre-processing and Analysis}
 Although pre-processing and the amount of training data will strongly affect detection accuracy for automatic glomerular analysis, most literature about glomerulus classification and detection mention little about these problems. For example, Kawazoe \etal\ \cite{kawazoe2018faster} proposed a method of using Faster R-CNN to detect glomeruli on periodic acid-Schiff (PAS), periodic acid-methenamine silver (PAM), Masson trichrome (MT) and Azan stained renal WSIs. In their methods, images are taken by $40 \times$ objective lens magnification and detection is conducted with downsampling equivalent to $5 \times$ objective lens magnification. Then the down-sampled images are sliding-window cropped by $1099 \times 1099$ pixels. For training data, images are cropped by the window centred on each annotated glomerulus and incomplete glomerular bounding boxes within the sliding window are ignored. Zhao \etal\ \cite{zhao2018dgdi} published a renal DIF dataset and explored several CNN methods to detect the glomeruli in renal DIF. For all the experiments, the authors used 12 times downsampling on the WSIs and cropped the resized images into patches for detection. Simon \etal\ \cite{simon2018multi} used local binary patterns (LBPs) image feature vector to train a support vector machine (SVM) model to classify glomeruli on light microscopy (LM) renal images. Their training images were extracted manually at a size of 576 $\times$ 576 pixels from the original WSIs. For test data, each WSI was fed into the SVM classifier by the sliding-window approach with a stride of 64 pixels. Window size was the same as the size of the training images ($576 \times 576$ pixels). Gallego \etal\ \cite{gallego2018glomerulus} proposed an AlexNet based CNN classifier to classify glomeruli for the PAS stained LM renal data. They also manually extracted glomerulus and non-glomerulus patches from the WSIs and resized all images to $227 \times 227$ pixels. In contrast to these works on renal WSI, here we perform extensive experiments about how various pre-processing methods and the amount of training data affect glomerulus detection performance on several state-of-the-art CNN models.

\subsection{CNN Based Object Detectors}\label{sec:CNN detector}

Since the CNN was first proposed for object detection task as Region-based CNN (R-CNN) by Girshick \etal\ \cite{girshick2016region}, various CNN-based models have produced impressive object detection performance. Several works have applied CNN methods to the computational pathology domain, such as cancer detection \cite{esteva2017dermatologist, hou2016patch, sun2017enhancing}, organ segmentation on CT and MRI images \cite{chartrand2017liver, okada2015abdominal} and cell classification \cite{chen2016deep, gao2017hep}. Modern CNN based object detectors can be roughly categorized into two categories: the two-stage detectors, such as Faster R-CNN \cite{ren2015faster}, Mask R-CNN \cite{he2017mask}, and R-FCN \cite{dai2016r}, and the one-stage detectors, such as YOLO \cite{redmon2016you}, YOLOv2 \cite{redmon2017yolo9000}, SSD \cite{liu2016ssd} and RetinaNet \cite{lin2017focal}. In the two-stage detector approach, the input image is first fed to a Region Proposal Network (RPN) to generate a sparse set of candidate boxes. These Region of Interests (RoIs) are then further classified and regressed by RoI-wise branches to generate the final prediction results \cite{dai2016r}. Different from the two-stage detectors, one-stage detectors do not generate region proposals. This significantly reduces running time at the price of reduction in accuracy. We employed four object detectors for our experiments: Faster R-CNN, R-FCN, Mask R-CNN and SSD.

\textbf{Faster R-CNN} \cite{ren2015faster} is an improved version of R-CNN \cite{girshick2016region} and Fast R-CNN \cite{Girshick2015}. Inspired by image classification, R-CNN directly applies a CNN based image classifier on a set of generated region proposals \cite{huang2017speed}. Although R-CNN improves the state-of-the-art detection accuracy, the proposal features are calculated multiple times which leads to large run time \cite{huang2017speed}. Fast R-CNN alleviates this problem by making all region proposal features share one-time generated feature extraction \cite{huang2017speed}. However, both R-CNN and Fast R-CNN depend on external proposal generators which then become the new bottleneck as everything apart from the regional proposal generator runs in the GPU. Faster R-CNN solves this problem by using a neural network called RPN to generate the candidate anchors and it is then able to be trained end-to-end.

\textbf{R-FCN} \cite{dai2016r} is proposed based on Faster R-CNN. R-FCN modifies the backbone network used for feature extraction in Faster R-CNN. This modification is crucial as a typical backbone network is trained for image classification problems which require the network to impose a translation invariant property. This property, however, is the opposite of the object detection problem which requires translation variance. To this end, R-FCN uses a position-sensitive convolution layer that is specifically trained to remove the translation invariant property.

\textbf{Mask R-CNN} \cite{he2017mask} is mainly targeted to address the instance segmentation problem. However, there are several improvements that allow Mask R-CNN to outperform Faster R-CNN. For instance, to perform accurate spatial quantization for feature extraction, RoI Align is used instead of RoI Pool. In addition, unlike Faster R-CNN, Mask RCNN uses a Feature Pyramid Network (FPN)\cite{lin2017feature} with ResNet as its backbone. FPN builds an in-network feature pyramid from a single-scale input by a top-down architecture with lateral connections \cite{lin2017feature}.

\textbf{SSD} \cite{liu2016ssd} is a single-stage object detector. SSD is similar to RPN, since both provide the detection results in one step. The difference is that whilst RPN provides object/non-object classification, SSD provides class-level classification. In other words, it directly classifies default anchor boxes’ classes and regresses their real bounding boxes. SSD combines predictions from multiple feature maps with different resolutions to handle various object sizes.

\section{ Experimental Setup and Protocols}
\label{ex}
\subsection{Data Collection}\label{sec:data collection}

\begin{figure}[htbp]
	\centering
	\includegraphics[width=0.4\textwidth,keepaspectratio]{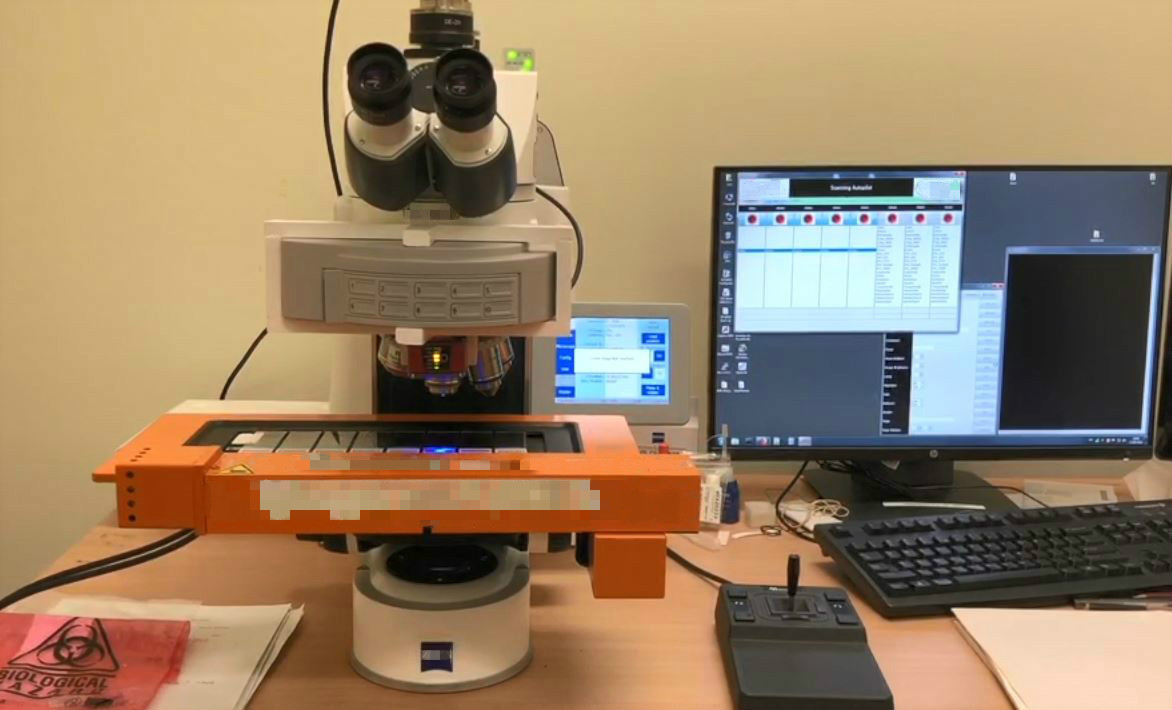}
	\caption{ The scanning system that digitized the renal glass slides.}
	\label{fig:microscope}
\end{figure}

Tissue samples from biopsies are digitized into TIF format using a M12 microscopy camera with the Sony IMX253 CMOS global shutter sensor at $40 \times$ objective lens magnification. Figure~\ref{fig:microscope} shows our scanning system that digitized the renal glass slides.
The system uses a two-stage scanning method that first creates a general view of the specimen using the low magnification objective~($5 \times$), and then this overview image is used as a location guide to scan the actual specimen using the high magnification objective~($ 40 \times$).

\begin{figure}[htbp]
	\centering
	\subfigure[Glomerular examples]{
		\begin{minipage}{0.5\textwidth}
			\centering
			\includegraphics[height=3cm,width=8cm]{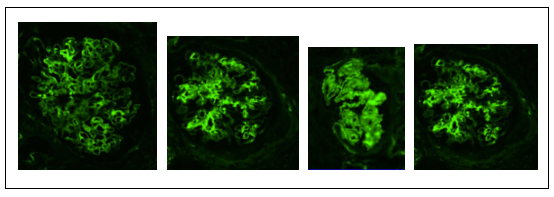}			
		\end{minipage}
	}
	\subfigure[Non-glomerular examples]{
		\begin{minipage}{0.5\textwidth}
			\centering
			\includegraphics[height=3cm,width=8cm]{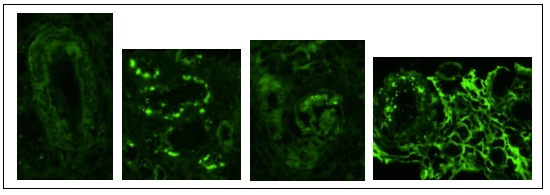}			
		\end{minipage}
	}
	\caption{Glomerulus and non-glomerulus examples.}
	\label{fig:glom example}
\end{figure}

The renal DIF dataset used in all experiments includes 230 WSIs collected from 30 patients. These WSIs have an average file size of 20 gigabytes and an average image size of $90,000 \times 72,000$ pixels. The antibodies for staining were: IgG, IgA, IgM, C3, Fib, C1q, Kappa and Lambda. Each WSI image was manually labelled. The labelling works were conducted by expert pathologists. Figure~\ref{fig:glom example} shows some glomerular and non-glomerular examples from our dataset. Compared to the generic object detection tasks, glomerulus detection on renal DIF images is more challenging. Renal DIF WSIs are enormous (90,000 $\times$ 72,000 pixels), but the size of glomeruli are relatively quite small, ranging from 4000 $\times$ 4000 pixels to 7000 $\times$ 7000 pixels. The glomerular staining intensity is highly variable as it relates to the different positivity grades of patients. In addition, due to different diseases, the patterns of glomeruli on renal DIF images can vary with conditions such as granular glomerular staining, linear glomerular deposit, and scanty glomerular immunostaining \cite{zhao2018dgdi}.

The raw renal DIF images generated from the camera were downsized and converted into JPEG images before being sent to the CNN detectors. Note that the downsized images were still very large. For example, after 12 times downsizing, the image was still  $8,000 \times 8,000$ pixels and could not be directly fed into the detection models. Thus, after resizing, the shrunk images were divided into overlapping patches of size of $1024 \times 1024$ pixels. The sliding-window method with stride of 256 pixels was used to perform the cropping. After getting the $1024 \times 1024$ patch-level detection result, it was mapped back to the WSIs to generate the final prediction result. Figure \ref{fig:wsi_detection} shows an example of the WSI detection result.

\subsection{Detector Setup}

The cropped patches were randomly divided into training set, validation set, and test set with a split ratio of 70\%, 10\% and 20\%, respectively. Note that, all data from the same patient was put into either the train or test set only. This patient-specific setting was required as a trained system should not have any data from an unseen new patient.

In our experiments, Faster R-CNN, Mask R-CNN and R-FCN used Resnet101~\cite{he2016deep} as the backbone network. SSD used Mobilenet v2~\cite{sandler2018mobilenetv2} as the backbone. All models were implemented using the Tensorflow framework. These models were pre-trained on the COCO dataset~\cite{lin2014microsoft} and then fine-tuned on our renal training set for 120,000 steps. Each model was trained and tested with the same pre-processing parameters. Faster R-CNN, Mask R-CNN and R-FCN used a batch size of 1 and the Stochastic Gradient Descent~(SGD) with momentum value of 0.9 as optimizer. SSD used a batch size of 32 and RMSprop~\cite{tieleman2012rmsprop} with momentum value of 0.9 as optimizer. The input image size for Faster R-CNN, Mask R-CNN and R-FCN was $1024 \times 1024$ pixels. The patches were further resized to $300 \times 300$ pixels for SSD in order to align with SSD pre-trained models. The learning rate was set at $10^{-3}$ with the weight decay regularization of $10^{-4}$. All experiments were performed on an NVIDIA Tesla V100 GPU.

\subsection{Evaluation Metric}\label{sec:Evaluation Metric}

For assessing experimental results, we use mAP as our main evaluation metric. The mAP calculates the area under the precision/recall curve. In addition, we further drill down into what the pre-processing and the amount of training data affect --- precision or recall. Equation (\ref{eq 1}) defines precision (P) and (\ref{eq 2}) defines recall (R), in terms of True Positive (TP), False Positive (FP) and False Negative (FN).

\begin{gather}
P = \dfrac{TP}{TP + FP} \label{eq 1}\\
R = \dfrac{TP}{TP + FN} \label{eq 2}
\end{gather}

\begin{figure}[htbp]
	\centering
	\includegraphics[width=0.3\textwidth,keepaspectratio]{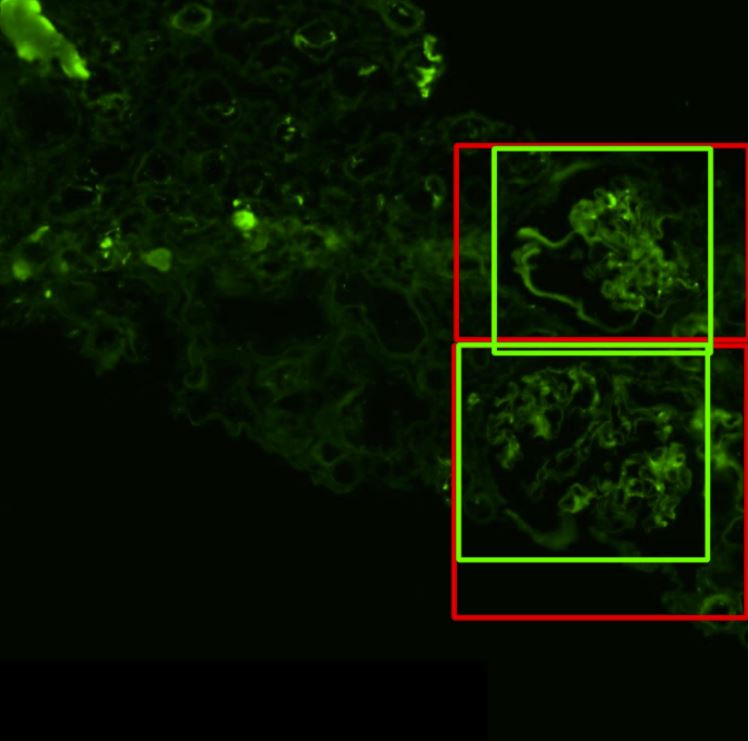}
	\caption{An example of the 8 times downsampled patch-level glomerulus detection result. The red boxes and green boxes are the ground truth and predicted bounding boxes, respectively. The two glomeruli do not have clear boundaries. It is hard to say that ground truth boxes are more accurate than predicted bounding boxes.}
	\label{fig:glomerulus detection example}
\end{figure}

Glomerulus detection is a challenging task as glomeruli vary in size, shape, and pattern. Sometimes glomeruli do not even have visible boundaries because their structures are damaged due to disease or the specimen preparation procedure. Therefore, even the annotation work is performed by pathologists, it is difficult to be 100\% sure where the accurate boundaries are located in DIF renal images. An example is shown in Figure \ref{fig:glomerulus detection example}. The red boxes are the ground truth boxes and the green boxes are the predicted boxes. It is difficult to determine whether the ground truth boxes are more accurate than the predicted bounding boxes. Furthermore, our primary concern is to localize the glomeruli to help the pathologists, rather than producing accurate boundaries. Thus, we focus on the mAP result at Intersection of Union (IoU) threshold of 0.5.

\section{Experimental Results and Analysis}
\label{re}
In order to explore the key settings relating to the training images which greatly affect the glomerulus detection result, we performed extensive experiments in four factors: downsampling, compression, training data amount, and different detection models.

\subsection{Experiments on Downsampling}
\begin{figure}[htbp]
	\centering
	\includegraphics[height=4cm,width=7.5cm]{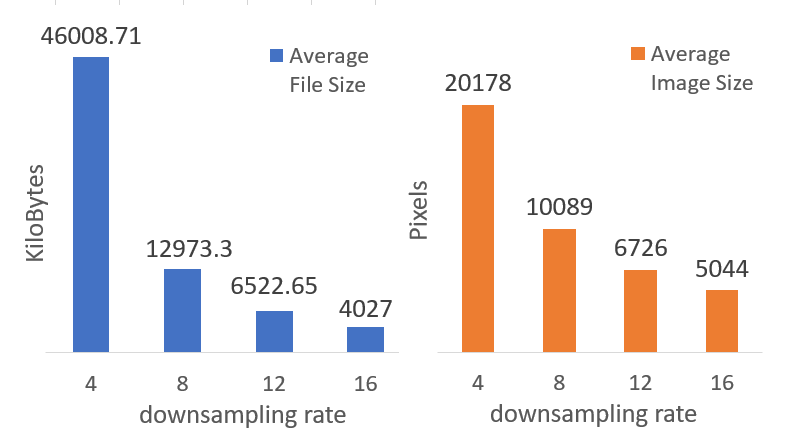}
	\caption{Average file size and image size per image against downsampling rate with no compression. Downsampling changes both input image's file size and image resolution.}
	\label{fig:shrink file size and image size}
\end{figure}

Downsampling and compression are two commonly used pre-processing methods to handle large images. The main difference between downsizing and compression is that downsizing reduces both the input image's file size and image size (resolution), but compression only reduces the input image's file size at the expense of some compression artefacts. As mentioned in Section \ref{sec:data collection}, the raw WSIs are extremely large and the resizing operation is always required. Thus, we want to evaluate to what extent the original uncompressed~(raw) TIF images can be resized whilst maintaining the detection accuracy. To study this, we downsize the original TIF images by 4, 8, 12 and 16. Then the downsized images are saved as JPEG files with no compression. Figure \ref{fig:shrink file size and image size} shows the average file size and image size per image at different downsampling rates. For example, when downsizing a TIF image by 4 and then saving it as JPEG, file size is decreased by 445 times and image resolution is only reduced by 4 from the original TIF file.

\begin{figure}[htbp]
	\includegraphics[height=6.5cm,width=9cm]{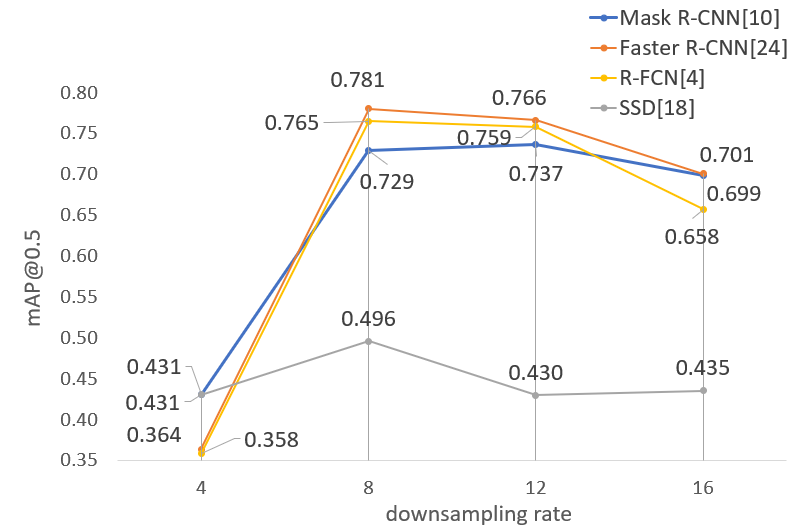}
	\caption{The mAP performance of four detectors trained and tested on images against downsampling rate.}
	\label{fig:shrink result}
\end{figure}

\begin{figure}[htbp]
	\includegraphics[height=6cm,width=9cm]{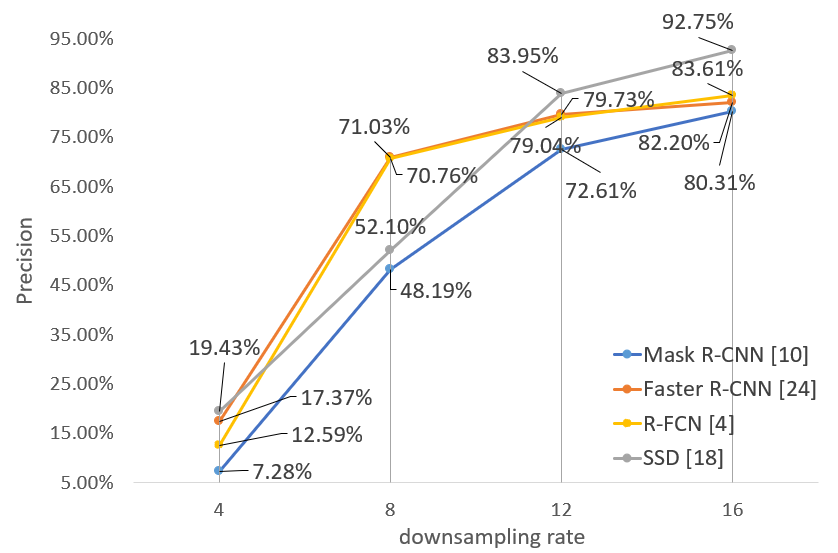}
	\caption{The precision performance of four detectors trained and tested against downsampling rate.}
	\label{fig:shrink result precision}
\end{figure}

\begin{figure}[htbp]
	\includegraphics[height=6cm,width=9cm]{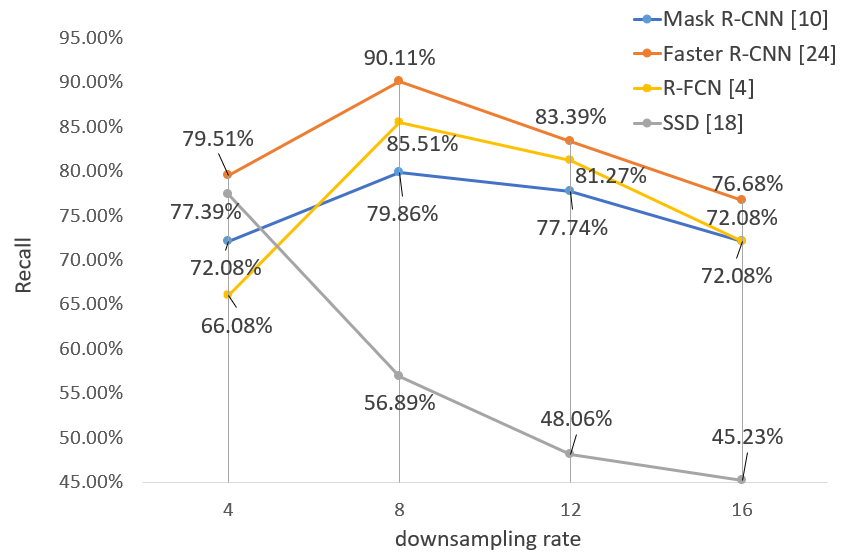}
	\caption{The recall performance of four detectors trained and tested against downsampling rate.}
	\label{fig:shrink result recall}
\end{figure}

If the downsampling is less than 4, the size of many glomeruli will be greater than the patch size (1024 $\times$ 1024 pixels). This will cause many cropped patches to only contain a sub-region of the glomeruli and impede model learning. Thus, we use downsampling rate at 4 as the minimum rate. Figures \ref{fig:shrink result},  \ref{fig:shrink result precision} and  \ref{fig:shrink result recall} show the mAP, precision and recall performance of CNN detectors trained and tested with images at different downsizing rates, respectively. Observing the experimental results, we find that apart from 8 times downsampling which actually increases accuracy, increasing the downsampling results in a drop in detection accuracy. At 8 times downsampling, all four detectors attain their best performance, especially for Faster R-CNN which achieves a mAP accuracy of 0.781.

At 4 times downsampling, all detectors have abysmal performance. This is because the glomerulus occupies most of the patch and there is little background seen in the training patches. The models are not provided with enough background information to learn the difference between glomeruli and background noise. Therefore they suffer from false positive problem leading to low precision as shown in Figure~\ref{fig:shrink result precision}. With increasing downsampling rate, all models' precision increases, since more background information is provided within the training set to avoid false positives. In contrast to the two-stage detectors, SSD has much lower detection accuracy. SSD's low performance is due to its high false negative rate (low recall) as shown in Figure \ref{fig:shrink result recall}.  

Due to the two-step cascaded classification and regression mechanism \cite{zhang2018single}, two-stage detectors are more robust for glomerulus detection than one-stage detectors. To sum up, for glomerulus detection, when the detector's input image size is set to $1024 \times 1024$ pixels, a downsampling rate of 8 times is optimal in terms of detection accuracy and small file size. Training a model using images with very low downsampling rate~(4 times), increases the false positive rate as not enough background is visible. On the other hand, significantly reducing the image size increases the false negative rate as the glomerulus becomes small and ambiguous.
These observations are corroborated by the findings in \cite{Marculescu2019} for general object detection task, which states that re-scaling the image to a lower resolution may produce better accuracy.

\subsection{Experiments on Compression}

\begin{figure}[htbp]
	\centering
	\includegraphics[height=4cm,width=8cm]{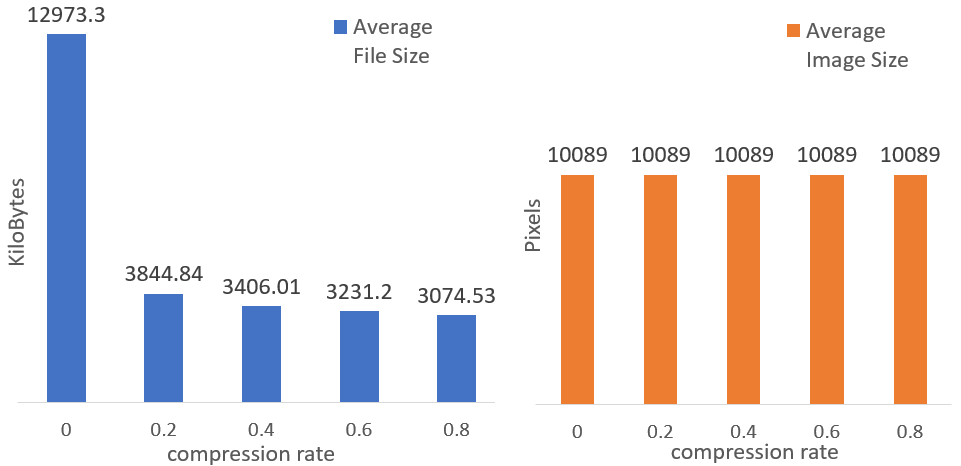}
	\caption{Average file size and image size per image at 8 times downsampling against compression rate. Compression only changes input image's file size and does not change its image resolution.}
	\label{fig:compress file size and image size}
\end{figure}

\begin{figure}[htbp]	
	\includegraphics[height=6cm,width=8cm]{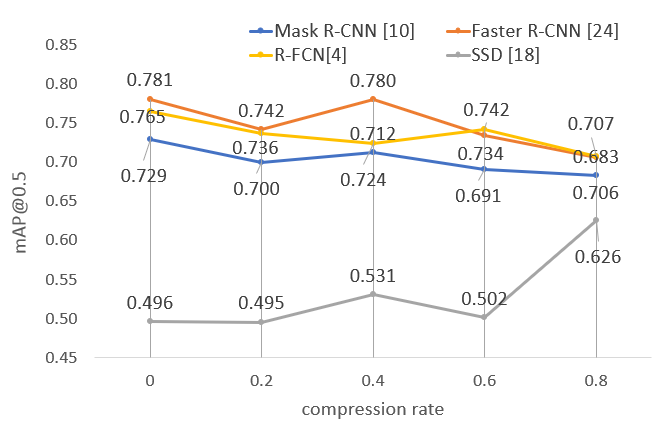}
	\caption{The mAP performance of four detectors trained and tested with images under 8 times downsizing against compression rate.}
	\label{fig:compress result}
\end{figure}

\begin{figure}[htbp]	
	\includegraphics[height=6cm,width=8cm]{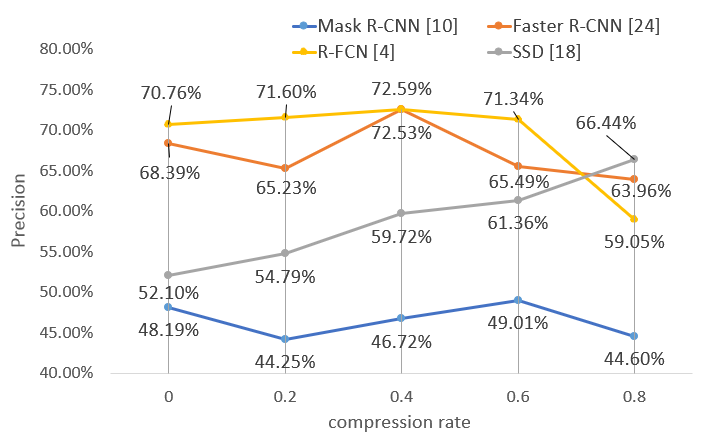}
	\caption{The precision performance of four detectors trained and tested with images under 8 times downsizing against compression rate.}
	\label{fig:compress result precision}
\end{figure}

\begin{figure}[htbp]	
	\includegraphics[height=6cm,width=8cm]{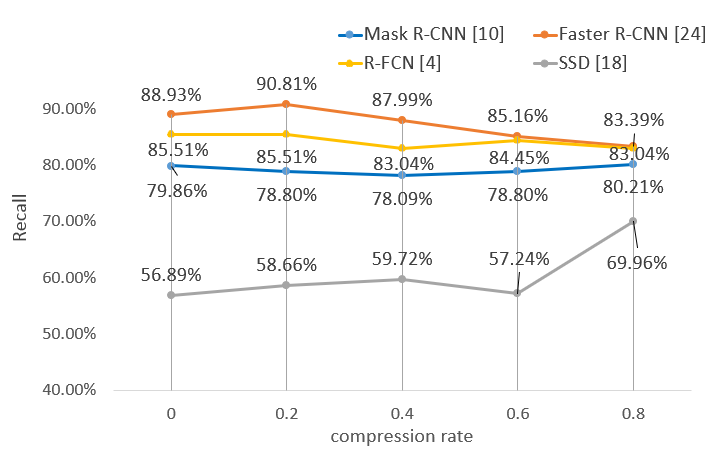}
	\caption{The recall performance of four detectors trained and tested with images under 8 times downsizing against compression rate.}
	\label{fig:compress result recall}
\end{figure}

Compression is also a good method to further reduce file size. All the images for the compression experiments are first downsampled to the optimal rate of 8 times. Then the resized images are compressed with JPEG compression rates of 0\%, 20\%, 40\%, 60\% and 80\%. Figure \ref{fig:compress file size and image size} shows the average file size and image size at different JPEG compression rates after the 8 times downsampling. Figures \ref{fig:compress result},  \ref{fig:compress result precision} and  \ref{fig:compress result recall} show the mAP, precision, and recall performance of the four detectors trained and tested on the downsized images with different compression rates.

From the results, we find that for two-stage detectors, except for a few exceptions, performance trends downwards when compression rate increases. In contrast to the two-stage detectors which have a significant accuracy drop at 80\% compression, SSD has an unexpected accuracy increase in accuracy. Due to its single-shot detection, SSD suffers from a strong false negative problem as shown in Figure \ref{fig:compress result recall}. We conjecture that with increasing compression rate, background noise becomes more random and dissimilar to the glomeruli appearance. This makes SSD can make better detection. For other small anomalies in the graph for the two-stage detectors, we conjecture similarly that JPEG compression artefacts may sometimes help differentiate the glomeruli from the background.

When compressing the downsized renal image at 40\%, file size drops from about 12.7 Megabytes (MB) to about 3.3 MB  which is roughly 4 times smaller. In addition, according to the results in Figure \ref{fig:compress result}, there is less than a 0.01 mAP accuracy decrease for both Faster R-CNN and Mask R-CNN. For SSD, its accuracy even increases by 0.035 mAP. Therefore, although pathology guidelines require raw images to be stored, our experimental results suggest that a suitable JPEG compression rate may help reduce file size with negligible cost to detection accuracy  --- at least for machine analysis, if not a human pathologist.

\subsection{Experiments on Training Data Size}
\begin{figure}[htbp]
	\centering
	\includegraphics[height=6cm,width=8cm]{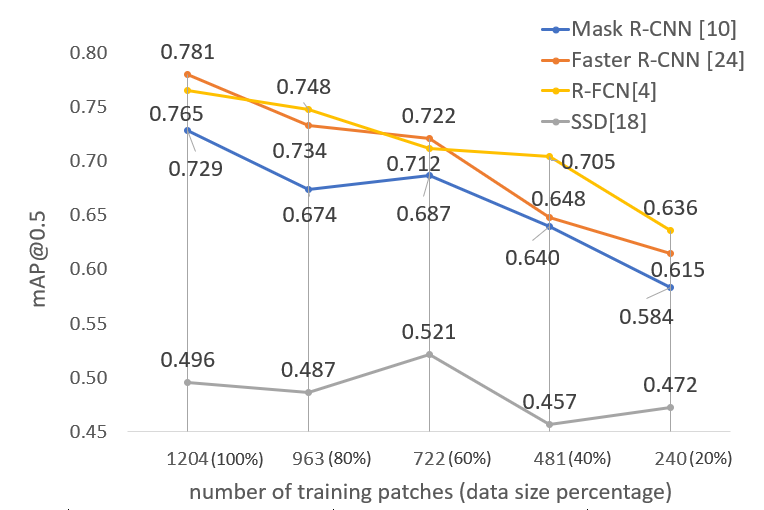}
	\caption{The mAP performance of four detectors trained and tested on 8 times downsized images with no JPEG compression against different training set sizes.}
	\label{fig:data size result}
\end{figure}

\begin{figure}[htbp]	
	\includegraphics[height=6cm,width=8cm]{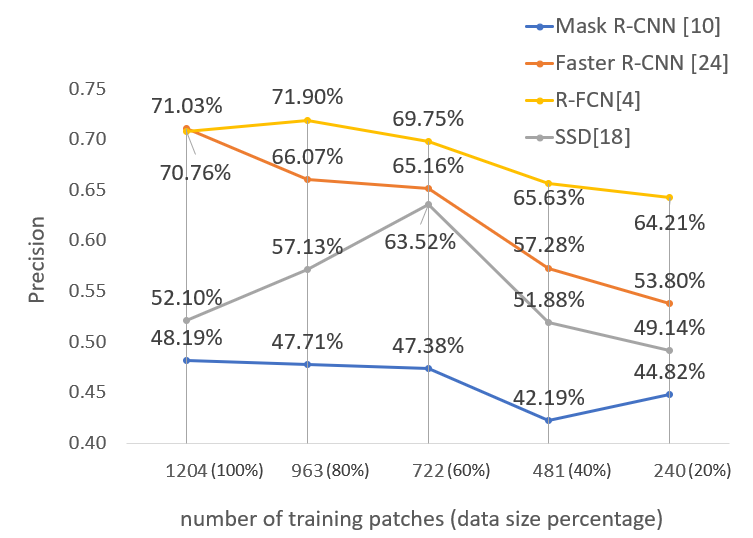}
	\caption{The precision performance of four detectors trained and tested on 8 times downsized images with no JPEG compression against different training set sizes.}
	\label{fig:data size precision}
\end{figure}

\begin{figure}[htbp]	
	\includegraphics[height=5cm,width=8cm]{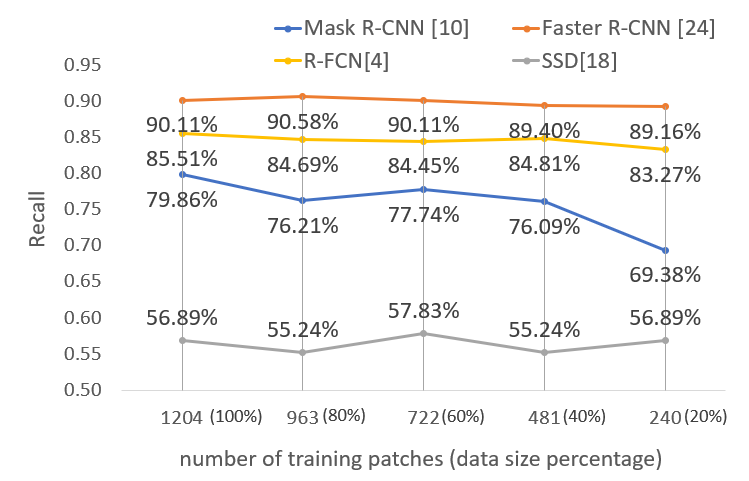}
	\caption{The recall performance of four detectors trained and tested on 8 times downsized images with no JPEG compression against different training set sizes.}
	\label{fig:data size recall}
\end{figure}

To explore how much data is required to train a model to accurately detect glomeruli on renal DIF images, we performed experiments with different sizes of training data. All training images are first downsampled 8 times and then saved as JPEG files with no compression. Figure \ref{fig:data size result}, Figure \ref{fig:data size precision} and Figure \ref{fig:data size recall} shows the mAP, precision and recall performance of the four detectors trained with different amounts of training data respectively. The number of WSIs used for the training set is 158 (1204 glomerular patches) from 20 patients. We hold back 64 WSIs from another 9 patients as the test set and 8 WSIs from 1 patient are used as the validation set. We progressively reduce the training set, while the validation and test sets are unchanged. This step is performed three times and the average performance is reported.

Observing the experimental results in Figure~\ref{fig:data size result}, we find that with reduction in training set size, the mAP performance drops. However, the negative trend is not linear and the mAP drop is much smaller than the reduction in the data size. For instance, when the training set is reduced from 80\% to 60\%, Faster R-CNN only suffers a 0.01 accuracy drop (from 0.734 to 0.722). According to Figure \ref{fig:data size precision} and Figure \ref{fig:data size recall}, the accuracy drop is mainly due to false positive instead of false negative. With the reducing of the training data amount, in most cases, two-stage detectors' precision drops. SSD presents a pulse at 60\% training data, we conjecture SSD's unstable performance is due to its single shot manner which makes it easier to be affected by background noise. In summary, lack of training data may cause a decrease on detection accuracy but the effect is not directly proportional. The performance reduction is mainly due to false positive. Fewer training data to some extent leads to worse precision but has little effect towards recall.

\subsection{Further Discussions on Different CNN Detectors}\label{sec:different CNN models}

By observing the performance of the selected four detectors in Figure \ref{fig:shrink result}, Figures \ref{fig:compress result} and \ref{fig:data size result}, we find that at an IoU threshold of 0.5, Faster R-CNN always gets the best performance followed by R-FCN, Mask R-CNN, and SSD. One reason why Mask R-CNN has lower performance than Faster R-CNN and R-FCN may be because the ground truth masks fed into Mask R-CNN have background noise. As mentioned in Section \ref{sec:Evaluation Metric}, for renal DIF images, glomeruli have high variable in size, shape and pattern and some of them even do not have visible boundaries. Thus, during manually labelling, we are only able to draw the bounding boxes on the glomeruli. For all Mask R-CNN experiments, we use ground truth bounding boxes as ground truth masks which hampers the Mask R-CNN performance and leads to high false positive (low precision). Therefore, although for generic object detection task, such as COCO dataset \cite{lin2014microsoft}, Mask R-CNN has good performance, in our scenario, Faster R-CNN is more recommendable.

\textbf{Detection speed} of different detectors are also analysed in Table \ref{tab:speed table}. Due to its single-shot approach, SSD suffers from false negative problem and has much lower detection accuracy compared to two-stage detectors, but it has the fastest detection speed. Within the three two-stage detectors, R-FCN has the fastest detection speed. The main difference between Faster R-CNN and R-FCN is the depth of the RoI sub-network. R-FCN extracts features from the final convolutional layer of ResNet101 network and uses position sensitive score maps and position sensitive RoI Pooling to get location information. Thus R-FCN has a shallower RoI sub-network than Faster R-CNN which helps to increase speed.

\begin{table}[htbp]
	\caption\small{Average detection speed of different detectors on the WSI patch (1024 $\times$ 1024 pixels).}
	
	\label{tab:speed table}	
	\begin{tabular}{|c|p{3.3cm}<{\centering}|}
		\hline
		Methods & Average Testing Time Per WSI Patch\\
		\hline
		Mask R-CNN~(ResNet101)~\cite{he2017mask} & 1036.17 ms\\
		\hline
		Faster R-CNN~(ResNet101)~\cite{ren2015faster} & 865.33 ms\\
		\hline
		R-FCN~(ResNet101)~\cite{dai2016r} & 810.21 ms\\
		\hline
		SSD~(MobileNet v2)~\cite{liu2016ssd} & 745.46 ms\\
		\hline
	\end{tabular}	
\end{table}
\section{Conclusion}
\label{conclu}
For general object detection, the most important elements are detection accuracy and speed. There are two extra key elements for glomerulus detection: storage space and data scarcity. Downsampling affects detection accuracy, speed, and storage space by changing image resolution, number of patches and file size. Compression can further alleviate the storage problem caused by WSIs --- often with negligible accuracy loss.

We have performed several experiments to find how the following factors affect glomerulus detection: downsampling rate, compression rate, amount of training data, and choice of detection models. For the best trade-off between performance, speed and storage, we conclude that using both resizing and compression to reduce WSI file size may have negligible effects on the final result, but can save a huge amount of storage space. At 8 times downsizing, Faster R-CNN can achieve 0.781 mAP at an IoU of 0.5 with file size 1616 times smaller than the original WSI. With both 40\% compression and 8 times downsizing, Faster R-CNN can achieve 0.780 mAP at an IoU of 0.5 with file size 6157 times smaller than the original WSI. Lack of training data will lead to decreased accuracy and increased false positives, but the effect is not directly proportional. Finally, when investigating different CNN models, since detection accuracy instead of speed is our primary concern, accurate two-stage detectors are preferred. Although Mask R-CNN and R-FCN add novel modifications to Faster R-CNN and achieve exciting performance on the COCO dataset, due to the nature of renal DIF data, these modifications do not lead to better performance on the glomerulus detection task.

\section*{Acknowledgements}
This research was funded by the Australian Government through the Australian Research Council and Sullivan Nicolaides Pathology under Linkage Project LP160101797. 

\small
\bibliographystyle{IEEEtranS}
\bibliography{reference}

\begin{thebibliography}{10}
\providecommand{\url}[1]{#1}
\csname url@samestyle\endcsname
\providecommand{\newblock}{\relax}
\providecommand{\bibinfo}[2]{#2}
\providecommand{\BIBentrySTDinterwordspacing}{\spaceskip=0pt\relax}
\providecommand{\BIBentryALTinterwordstretchfactor}{4}
\providecommand{\BIBentryALTinterwordspacing}{\spaceskip=\fontdimen2\font plus
\BIBentryALTinterwordstretchfactor\fontdimen3\font minus
  \fontdimen4\font\relax}
\providecommand{\BIBforeignlanguage}[2]{{%
\expandafter\ifx\csname l@#1\endcsname\relax
\typeout{** WARNING: IEEEtranS.bst: No hyphenation pattern has been}%
\typeout{** loaded for the language `#1'. Using the pattern for}%
\typeout{** the default language instead.}%
\else
\language=\csname l@#1\endcsname
\fi
#2}}
\providecommand{\BIBdecl}{\relax}
\BIBdecl

\bibitem{agarwal2013basics}
S.~Agarwal, S.~Sethi, and A.~Dinda, ``Basics of kidney biopsy: A nephrologist's
  perspective,'' \emph{Indian journal of nephrology}, vol.~23, no.~4, p. 243,
  2013.

\bibitem{chartrand2017liver}
G.~Chartrand, T.~Cresson, R.~Chav, A.~Gotra, A.~Tang, and J.~A. De~Guise,
  ``Liver segmentation on ct and mr using laplacian mesh optimization,''
  \emph{IEEE Transactions on Biomedical Engineering}, vol.~64, no.~9, pp.
  2110--2121, 2017.

\bibitem{chen2016deep}
C.~L. Chen, A.~Mahjoubfar, L.-C. Tai, I.~K. Blaby, A.~Huang, K.~R. Niazi, and
  B.~Jalali, ``Deep learning in label-free cell classification,''
  \emph{Scientific reports}, vol.~6, p. 21471, 2016.

\bibitem{dai2016r}
J.~Dai, Y.~Li, K.~He, and J.~Sun, ``R-fcn: Object detection via region-based
  fully convolutional networks,'' in \emph{Advances in neural information
  processing systems}, 2016, pp. 379--387.

\bibitem{esteva2017dermatologist}
A.~Esteva, B.~Kuprel, R.~A. Novoa, J.~Ko, S.~M. Swetter, H.~M. Blau, and
  S.~Thrun, ``Dermatologist-level classification of skin cancer with deep
  neural networks,'' \emph{Nature}, vol. 542, no. 7639, p. 115, 2017.

\bibitem{gallego2018glomerulus}
J.~Gallego, A.~Pedraza, S.~Lopez, G.~Steiner, L.~Gonzalez, A.~Laurinavicius,
  and G.~Bueno, ``Glomerulus classification and detection based on
  convolutional neural networks,'' \emph{Journal of Imaging}, vol.~4, no.~1,
  p.~20, 2018.

\bibitem{gao2017hep}
Z.~Gao, L.~Wang, L.~Zhou, and J.~Zhang, ``Hep-2 cell image classification with
  deep convolutional neural networks,'' \emph{IEEE journal of biomedical and
  health informatics}, vol.~21, no.~2, pp. 416--428, 2017.

\bibitem{Girshick2015}
R.~Girshick, ``Fast r-cnn,'' in \emph{Proceedings of the IEEE international
  conference on computer vision}, 2015, pp. 1440--1448.

\bibitem{girshick2016region}
R.~Girshick, J.~Donahue, T.~Darrell, and J.~Malik, ``Region-based convolutional
  networks for accurate object detection and segmentation,'' \emph{IEEE
  transactions on pattern analysis and machine intelligence}, vol.~38, no.~1,
  pp. 142--158, 2016.

\bibitem{he2017mask}
K.~He, G.~Gkioxari, P.~Doll{\'a}r, and R.~Girshick, ``Mask r-cnn,'' in
  \emph{Proceedings of the IEEE international conference on computer vision},
  2017, pp. 2961--2969.

\bibitem{he2016deep}
K.~He, X.~Zhang, S.~Ren, and J.~Sun, ``Deep residual learning for image
  recognition,'' in \emph{Proceedings of the IEEE conference on computer vision
  and pattern recognition}, 2016, pp. 770--778.

\bibitem{hou2016patch}
L.~Hou, D.~Samaras, T.~M. Kurc, Y.~Gao, J.~E. Davis, and J.~H. Saltz,
  ``Patch-based convolutional neural network for whole slide tissue image
  classification,'' in \emph{Proceedings of the IEEE Conference on Computer
  Vision and Pattern Recognition}, 2016, pp. 2424--2433.

\bibitem{huang2017speed}
J.~Huang, V.~Rathod, C.~Sun, M.~Zhu, A.~Korattikara, A.~Fathi, I.~Fischer,
  Z.~Wojna, Y.~Song, S.~Guadarrama \emph{et~al.}, ``Speed/accuracy trade-offs
  for modern convolutional object detectors,'' in \emph{Proceedings of the IEEE
  conference on computer vision and pattern recognition}, 2017, pp. 7310--7311.

\bibitem{kawazoe2018faster}
Y.~Kawazoe, K.~Shimamoto, R.~Yamaguchi, Y.~Shintani-Domoto, H.~Uozaki,
  M.~Fukayama, and K.~Ohe, ``Faster r-cnn-based glomerular detection in
  multistained human whole slide images,'' \emph{Journal of Imaging}, vol.~4,
  no.~7, p.~91, 2018.

\bibitem{lin2017feature}
T.-Y. Lin, P.~Doll{\'a}r, R.~Girshick, K.~He, B.~Hariharan, and S.~Belongie,
  ``Feature pyramid networks for object detection,'' in \emph{Proceedings of
  the IEEE Conference on Computer Vision and Pattern Recognition}, 2017, pp.
  2117--2125.

\bibitem{lin2017focal}
T.-Y. Lin, P.~Goyal, R.~Girshick, K.~He, and P.~Doll{\'a}r, ``Focal loss for
  dense object detection,'' in \emph{Proceedings of the IEEE international
  conference on computer vision}, 2017, pp. 2980--2988.

\bibitem{lin2014microsoft}
T.-Y. Lin, M.~Maire, S.~Belongie, J.~Hays, P.~Perona, D.~Ramanan,
  P.~Doll{\'a}r, and C.~L. Zitnick, ``Microsoft coco: Common objects in
  context,'' in \emph{European conference on computer vision}.\hskip 1em plus
  0.5em minus 0.4em\relax Springer, 2014, pp. 740--755.

\bibitem{liu2016ssd}
W.~Liu, D.~Anguelov, D.~Erhan, C.~Szegedy, S.~Reed, C.-Y. Fu, and A.~C. Berg,
  ``Ssd: Single shot multibox detector,'' in \emph{European conference on
  computer vision}.\hskip 1em plus 0.5em minus 0.4em\relax Springer, 2016, pp.
  21--37.

\bibitem{Marculescu2019}
T.-W. C. R. D.~D. Marculescu, ``Adascale: Towards real-time video object
  detection using adaptive scaling,'' in \emph{Conference on System and Machine
  Learning (SysML)}, 2019.

\bibitem{molne2005immunoperoxidase}
J.~M{\"o}lne, M.~E. Breimer, and C.~T. Svalander, ``Immunoperoxidase versus
  immunofluorescence in the assessment of human renal biopsies,''
  \emph{American journal of kidney diseases}, vol.~45, no.~4, pp. 674--683,
  2005.

\bibitem{okada2015abdominal}
T.~Okada, M.~G. Linguraru, M.~Hori, R.~M. Summers, N.~Tomiyama, and Y.~Sato,
  ``Abdominal multi-organ segmentation from ct images using conditional
  shape--location and unsupervised intensity priors,'' \emph{Medical image
  analysis}, vol.~26, no.~1, pp. 1--18, 2015.

\bibitem{redmon2016you}
J.~Redmon, S.~Divvala, R.~Girshick, and A.~Farhadi, ``You only look once:
  Unified, real-time object detection,'' in \emph{Proceedings of the IEEE
  conference on computer vision and pattern recognition}, 2016, pp. 779--788.

\bibitem{redmon2017yolo9000}
J.~Redmon and A.~Farhadi, ``Yolo9000: better, faster, stronger,'' in
  \emph{Proceedings of the IEEE conference on computer vision and pattern
  recognition}, 2017, pp. 7263--7271.

\bibitem{ren2015faster}
S.~Ren, K.~He, R.~Girshick, and J.~Sun, ``Faster r-cnn: Towards real-time
  object detection with region proposal networks,'' in \emph{Advances in neural
  information processing systems}, 2015, pp. 91--99.

\bibitem{samak2015optimization}
A.~Samak, A.~Wiliem, P.~Hobson, M.~Walsh, T.~Ditchmen, A.~Troskie,
  S.~Barksdale, R.~Edwards, A.~Jennings, and B.~C. Lovell, ``An optimization
  approach to scanning skin direct immunofluorescence specimens,'' in
  \emph{2015 International Conference on Digital Image Computing: Techniques
  and Applications (DICTA)}.\hskip 1em plus 0.5em minus 0.4em\relax IEEE, 2015,
  pp. 1--8.

\bibitem{sandler2018mobilenetv2}
M.~Sandler, A.~Howard, M.~Zhu, A.~Zhmoginov, and L.-C. Chen, ``Mobilenetv2:
  Inverted residuals and linear bottlenecks,'' in \emph{Proceedings of the IEEE
  Conference on Computer Vision and Pattern Recognition}, 2018, pp. 4510--4520.

\bibitem{simon2018multi}
O.~Simon, R.~Yacoub, S.~Jain, J.~E. Tomaszewski, and P.~Sarder, ``Multi-radial
  lbp features as a tool for rapid glomerular detection and assessment in whole
  slide histopathology images,'' \emph{Scientific reports}, vol.~8, no.~1, p.
  2032, 2018.

\bibitem{sun2017enhancing}
W.~Sun, T.-L.~B. Tseng, J.~Zhang, and W.~Qian, ``Enhancing deep convolutional
  neural network scheme for breast cancer diagnosis with unlabeled data,''
  \emph{Computerized Medical Imaging and Graphics}, vol.~57, pp. 4--9, 2017.

\bibitem{tieleman2012rmsprop}
T.~Tieleman and G.~Hinton, ``Rmsprop: Divide the gradient by a running average
  of its recent magnitude. coursera: Neural networks for machine learning,''
  \emph{COURSERA Neural Networks Mach. Learn}, 2012.

\bibitem{zhang2018single}
S.~Zhang, L.~Wen, X.~Bian, Z.~Lei, and S.~Z. Li, ``Single-shot refinement
  neural network for object detection,'' in \emph{Proceedings of the IEEE
  Conference on Computer Vision and Pattern Recognition}, 2018, pp. 4203--4212.

\bibitem{zhao2018dgdi}
K.~Zhao, Y.~J.~J. Tang, T.~Zhang, J.~Carvajal, D.~F. Smith, A.~Wiliem,
  P.~Hobson, A.~Jennings, and B.~C. Lovell, ``Dgdi: A dataset for detecting
  glomeruli on renal direct immunofluorescence,'' in \emph{2018 Digital Image
  Computing: Techniques and Applications (DICTA)}.\hskip 1em plus 0.5em minus
  0.4em\relax IEEE, 2018, pp. 1--7.

\end{thebibliography}

\end{document}